\documentclass[11pt]{article}
\usepackage{acl2015}
\usepackage{times}
\usepackage{url}
\usepackage{latexsym}
\usepackage{graphicx}
\usepackage{framed,enumitem} 

\title{Evaluating NLP Systems On a Novel Cloze Task: Judging the Plausibility of Possible Fillers in Instructional Texts}

\author{Zizhao Hu \\
  \small{University of Southern California}\\
  \And
  Ravikiran Chanumolu \\
  \small{University of Southern California}\\
  \And
  Xingyu Lin \\
  \small{University of Southern California}\\
  \AND
  Nayela Ayaz \\
  \small{University of Southern California}\\
  \And
  Vincent Chi \\
  \small{University of Southern California}\\
  \\}
\date{}

\begin{document}

\maketitle

\begin{abstract}
    Cloze task is a widely used task to evaluate an NLP system's language understanding ability. However, most of the existing cloze tasks only require NLP systems to give the relative best prediction for each input data sample, rather than the absolute quality of all possible predictions, in a consistent way across the input domain. Thus a new task is proposed: predicting if a filler word in a cloze task is a good, neutral, or bad candidate. Complicated versions can be extended to predicting more discrete classes or continuous scores. We focus on subtask A in Semeval 2022 task 7, explored some possible architectures to solve this new task, provided a detailed comparison of them, and proposed an ensemble method to improve traditional models in this new task.
    \end{abstract}

\section{Introduction \& Related Work}  

    WikiHow is the largest and the most trusted how-to website. It has more than 300,000 articles and over 2.5 million registered users. The primary goal of the articles is to help users achieve  a specific goal like how to cook a Thanksgiving turkey or how to deal with an online stalker. The instructional articles frequently undergo revisions to  correct grammatical errors, elaborate on the underspecified text and remove ambiguities. These clarifications are necessary to ensure the best user experience. The goal of the SemEval-2022 Task 7  is to evaluate the ability of NLP systems to automatically distinguish between plausible and implausible clarifications of instructional text. The task is formulated as a cloze task, in which clarifications are presented as possible fillers in a masked sentence. For each (sentence, filler) pair the system must find if the clarification is plausible, implausible, or neutral.
    
    TF-idf, Glove, and Word2Vec are widely used non-contextual embedding techniques, and such embeddings containing useful information on word-level without contextual information can be used to solve various NLP tasks.
    
    However, recent discoveries in contextualized embeddings and language models have shown their dominance. The current best architecture for solving language understanding tasks is based on fine-tuning a large pre-trained language model on the downstream task \cite{radford2018improving}. Pre-trained language models such as BERT \cite{devlin2018bert}, ELMo \cite{peters-etal-2018-deep}, OpenAI GPT \cite{radford2018improving} have shown state-of-the-art performance on multiple language understanding tasks. And evidence of such models' ability to have commonsense knowledge \cite{cui2020does} gives the empirical basis to use such models for the cloze task, which is a sub-task that requires commonsense in the language understanding domain.

    Masked language modeling(MLM) task, a self-supervised, cloze task format pretraining objective, is widely used in the pre-training for transformer based language models. And due to its training nature, the pre-trained models with such objectives without fine-tuning can be directly applied to solve a general cloze task. Further fine-tuning on the training dataset can be conducted in the same manner as pre-training, to possibly improve the performance.
    
    The ability of the NLP system to give consistent judging(scoring) across the whole input domain has been exploited in the domain of Out-of-distribution(OOD) detection. There is proof that a well-designed scoring system can perform well in OOD detection. Such scoring system can be directly built upon the softmax output of the classification task) \cite{hendrycks2016baseline} or distance based \cite{chang2017weighted} (Training confidence calibrated classifiers for detecting out-of-distribution samples \cite{lee2017training}). Inspired by these discoveries, we designed our scoring systems

\section{Methods}

\subsection{Data}

\begin{figure}[!ht]
    \centering
    \includegraphics[width=5cm]{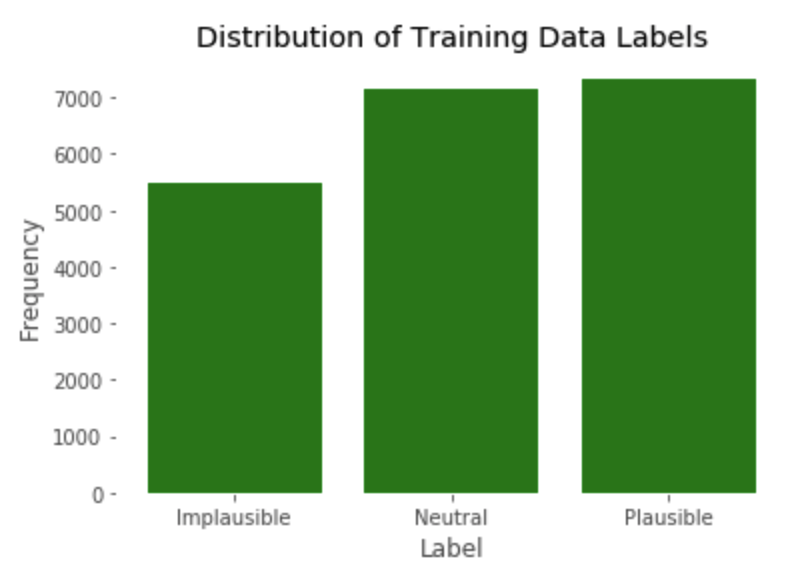}
    \includegraphics[width=5cm]{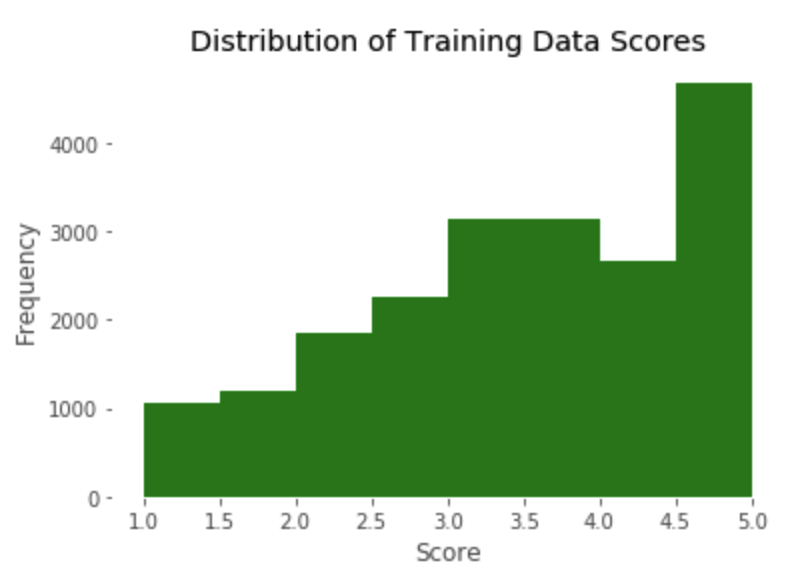}
    \caption{Training Data Distributions}
    \label{fig:galaxy}
\end{figure}

We took the data and labels from Semeval 2022 task 7. The data consists of the article title, the subheading, a masked sentence with its previous and next instructions, and the possible fillers with corresponding labels and ratings that predict the plausibility score. The label has 3 classes, positive (plausible), neutral, and negative (implausible), each indicating the quality of each candidate word. Numerical plausibility scores from 1-5 are also given for each training data example, and we can see the distribution in figure \ref{fig:galaxy}.

Following methods are used to preprocess the input texts:

\begin{enumerate}
\item{\textbf{Preprocessing Method 1}:
Concatenating titles, subheading, previous instruction, masked sentence, next instruction into one long sentence. }
\item{\textbf{Preprocessing Method 2}:
 Concatenating previous instruction, masked sentence, next instruction into one long sentence. }
\item{\textbf{Preprocessing Method 3}:
Only use the masked sentence.
}
\item{\textbf{Additional Preprocessing for MLM}:
Remove the first word if the filling word is a two-word pair: 'My book', 'The table'. Since the model can predict one masked word at a time.
}
\end{enumerate}

\begin{figure}[!ht]
    \centering
    \includegraphics[width=0.5\textwidth,height=4cm]{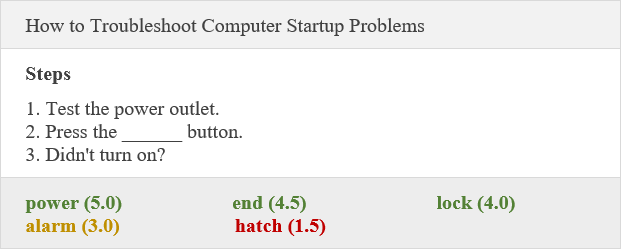}
    \caption{Sample Training Data with scores}
    \vspace{0.5cm}
    \includegraphics[width=0.5\textwidth,height=4cm]{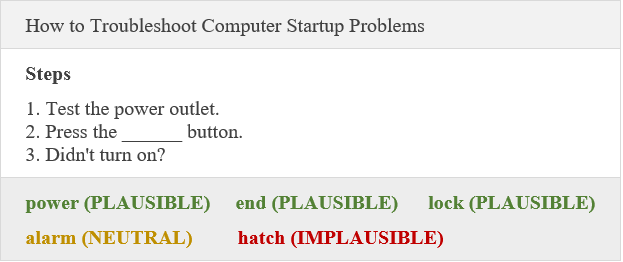}
    \caption{Sample Training Data with labels}
    \label{fig:sampledata2}
\end{figure}

\subsection{High Level Approaches}
We define a set of variables and functions $\{X, X', y, s, l, f_{emb}, f_{score}, f_{cls}, f_{reg}\}$. Here $\{x_1, x_2, ... x_n\}\in X$ is a set of the masked input text. For each masked input text $x_n$, it has 5 candidate filling words $\{y_{n_i}, i = 1,2,...5\} $. Each filling word has a corresponding label $l_{n_i} \in \{positive(5), neutral(3), negative(1)\}$ and a corresponding score $ s_{n_i} \in [1, 5]$. $\{x'_{n_i}, i = 1,2,...5\} \in X'$ is the set of filled input text. $f_{emb}$ is a non-contextual embedding method or contextual embedding taken from a language model. $f_{score}$ is a function that translates a filling word probability into a score, we'll discuss this in detail in section 3.1.2. $f_{cls}$ is a multi-class classifier. $f_{reg}$ is a regression model. 
\subsubsection{Approach 1}
Use embeddings to train a classifier $f_{cls}(f_{emb}(x'_{n_i}), l_{n_i})$. This approach corresponds to Experiment sections 3.1.1.
\subsubsection{Approach 2} 
Use a score to train a classifier$f_{cls}(f_{prob}(P(y_{n_i}|context)), l_{n_i})$, here the context is $x_n$ for MLM and neighboring words for N-gram models. This approach corresponds to Experiment sections 3.1.2 and 3.1.2.i.
\subsubsection{Approach 3} 
Use a score to train a regression model $f_{reg}(f_{prob}(P(y_{n_i}|context)), l_{n_i})$. This approach corresponds to Experiment sections 3.1.3 and 3.1.4.ii. The rule to decide labels is discussed in detail in the corresponding sections. 
\subsubsection{Evaluation Methods} 
We use overall accuracy and class-wise fscore to measure the classification performance.
In addition, we also use Spearman's rank correlation coefficient(Spearman rank) to assess monotonic relationships between the true scores and predicted labels. Since the labels have ordinal properties, we translate them to scores to make this evaluation method possible. the possible values ranging from -1 to 1. With 1 indicating the strongest relationship, -1 the opposite.


\subsection{Language Models}
    In this section we describe the different transformer models used in this work.

    \renewcommand{\theenumi}{\roman{enumi}}%
    \begin{enumerate}
    \item{\textbf{Bert:} Bert \cite{devlin2018bert} is a multi-layer transformer encoder model which achieved state-of-the-art performance on a wide variety of NLP tasks[]. The key technical innovation in Bert is using bidirectional self-attention to learn text representations.  The paper’s results show that a language model which is bi-directionally trained can have a deeper sense of language context than single-direction language models. Bert is therefore an ideal LM candidate for our task as it can score the correctness of each (sentence, filler) pair by jointly conditioning on both the right and left context. 
    }
    
    \item{\textbf{XLNet:}} XLNet \cite{yang2019xlnet} is a generalized autoregressive language model that is trained by maximizing log-likelihood over all the possible permutations of words in a sentence,  thereby getting bidirectional context information. XLNet is a good candidate for our task because such a model learns to get info from all positions in a sentence to determine the likelihood of a word at a certain position, which makes it a good candidate for our task, as plausibility depends on the entire sequence. 
    
    \item{\textbf{Electra:} Electra replaces the MLM training objective of BERT with Replaced Token Detection (RTD) \cite{clark2020electra}. The Electra model has two components a generator model and a discriminator model. The discriminator model takes as input a corrupted sequence. For each token in the sequence, it outputs a probability score based on whether the token is corrupted or not. The generator model is used to create the corrupted sequence. This is achieved by masking an input sentence and using a small Bert model, trained through MLM to replace the mask with the most likely token. This formulation allows the decoder to learn from every token in the input sequence. In contrast, with the MLM objective, Bert only learns from 15\% of the tokens. After pretraining, the generator model is discarded and the discriminator model is used for fine-tuning the downstream tasks.}
    \end{enumerate}

\section{Experiments}
\subsection{Main Experiments}
Here we tested several model architectures:

\subsubsection{Classification with sentence embeddings}
In this approach, we convert each (sentence, filler) pair to a fixed size representation and train a simple classification model over them. To obtain the sentence representations from word embeddings we use the following approaches: When using Glove we use the simple average of word embeddings to obtain sentence representation. In the case of transformer models, we take the average of the word embeddings obtained from the last layer. Further, we use the embedding of the first word-piece token as the word embedding. We tried other variations which included averaging and concatenating the embeddings of different layers in the transformer. But this did not result in performance improvement. We omit these variations in the results section due to limited space. Below we enumerate different models using this approach. For each model, we show the embedding method followed by the classifier.
    
    \begin{enumerate}
        \item[(1)]{glove + Logistic Regression}
        \item[(2)]{BERT + Logistic Regression}
        \item[(3)]{XLNet + Logistic Regression}
        \item[(4)]{Electra + Logistic Regression}
    \end{enumerate}
    
    

\subsubsection{Classification with MLM scores}
In this approach, we take advantage of the similarity between our task and the MLM task and use the output logits for the whole vocabulary to train a simple classification model to map the logits to classification labels. We first translate the logits to scores(will discuss below). Then a classification model is used to map the score to predicted labels. Here are some ways to translate the logits to scores:
\begin{enumerate}
    \item{\textbf{Logits}: Use the logits directly as scores.}
     \begin{enumerate}
        \item[(9)]{BERT + Naive Bayes Classifier}
    \end{enumerate}
    \item{\textbf{Soft-max}: Add a soft-max layer to the MLM and use the soft-max probability as scores.}
    \begin{enumerate}
        \item[(10)]{BERT + Naive Bayes Classifier}
    \end{enumerate}

    \item{\textbf{Soft-max + Top-1 Embedding Similarity}: Add a soft-max layer to the MLM, take the word with the highest probability. Filling the word into the masked sentence and acquiring the average embedding according to 3.2.1.i. Fill the candidate word into the masked sentence and acquire the average embedding. Use the cosine similarity between these two embeddings as the score. 
    }
    \begin{enumerate}
        \item[(11)]{BERT + Naive Bayes Classifier}
    \end{enumerate}
     \item{\textbf{Soft-max + Weighted Top-5 Embedding Similarity}: The same way to acquire the cosine similarity score. But we choose the top 5 highest soft-max probability words and use the sum of the 5 cosine similarity score weighted by the corresponding soft-max probabilities as the score.
    }
    \begin{enumerate}
        \item[(12)]{BERT + Naive Bayes Classifier}
    \end{enumerate}
    \item{\textbf{Soft-max + Maximum Top-5 Embedding Similarity}: The same way to acquire the cosine similarity score. But we choose the top 5 highest soft-max probability words and use the largest cosine similarity as the score. 
    }
    \begin{enumerate}
        \item[(13)]{BERT + Naive Bayes Classifier}
    \end{enumerate}
    
\end{enumerate}

\subsubsection{Regression with MLM scores}
For a regression model, we use the values 1, 3, 5 for negative, neutral, and positive labels as the dependent variable. Then we set 2 thresholds on the dependent values, above which neutral and positive are predicted respectively otherwise negative. The thresholds are set to values that will predict the same amount of labels for each class as there are in the training set. The scores are acquired in the same way as mentioned in the previous section:
 \begin{enumerate}
    \item{\textbf{Logits}:}
        \begin{enumerate}
            \item[(14)]{BERT + Linear Regression}
        \end{enumerate}
    \item{\textbf{Soft-max}: }
        \begin{enumerate}
            \item[(15)]{BERT + Linear Regression}
        \end{enumerate}
 \end{enumerate}
 
\begin{table*}[h]
\begin{center}
\begin{tabular}{|c|c|c|c|c|c|} 
 \hline
    \textbf{Method} & \textbf{accuracy} & \textbf{fscore0} & \textbf{fscore1} & \textbf{fscore2} & \textbf{spearman rank} \\ 
 \hline\hline
    \multicolumn{6}{|l|}{Task Baseline} \\
    \hline
    tf-idf + NB & 0.38 & 0.01 & 0.15 & 0.56 & 0.024  \\ 
   
    \hline
    \multicolumn{6}{|l|}{Classification with sentence embeddings} \\
    \hline
    glove + LR & 0.34 & 0.09 & 0.22 & 0.52 & 0.098  \\ 
    BERT$_{base}$ + LR & 0.38 & 0.33 & 0.23 & 0.50 & 0.074 \\ 
    XLNet + LR & 0.36 & 0.29 & 0.25 & 0.46 & 0.019 \\
    Electra + LR & 0.40 & 0.33 & 0.25 & 0.53 & 0.150 \\
    \hline
    \multicolumn{6}{|l|}{Classification with MLM scores} \\
    \hline
    BERT$_{base}$ + NB & 0.45 & 0.59 & 0.01 & 0.26 & 0.244  \\ 
    BERT$_{base+adapt}$ + NB & 0.45 & 0.59 & 0.02 & 0.27 &  0.251 \ \\
    BERT$_{large}$ + NB & 0.46 & \textbf{0.60} & 0.02 & 0.28 & 0.288  \\ 
    XLNet + NB & 0.44 & 0.04 & 0.24 & 0.60 & 0.095 \\
    \hline
    \multicolumn{6}{|l|}{Classification with RTD scores} \\
    \hline
     Electra + NB & 0.39 & 0.01 & 0.25 & 0.58 & 0.088 \\
    \hline
    \multicolumn{6}{|l|}{Regression with MLM scores} \\
    \hline
    BERT$_{base}$ + LinearReg & 0.47 & 0.49 & 0.27 & 0.58 & 0.350 \\
    BERT$_{large}$ + LinearReg & \textbf{0.48} & 0.50 & \textbf{0.28} & \textbf{0.60} & \textbf{0.398}  \\ 
    \hline
    \multicolumn{6}{|l|}{Using N-gram probabilities} \\
    \hline
    N-gram + NB & 0.38 & 0.00 & 0.25 & 0.58 & 0.107  \\ 
    N-gram + LinearReg & 0.34 & 0.21 & 0.28 & 0.47 & 0.109  \\ 
    \hline
    \multicolumn{6}{|l|}{Pretraining with version distinction task} \\
    \hline
    Electra + NB & 0.36 & 0.18 & 0.20 & 0.52 & 0.164  \\

\hline
\end{tabular}
\label{table:T1}
\caption{Results}
{In this table we show performance of models listed under section 3.1 - Main Experiments. Note: All transformer models are base models unless explicitly mentioned.}
\end{center}
\end{table*}

\begin{table*}[h]
\begin{center}
\begin{tabular}{|c|c|c|c|c|c|} 
 \hline
    \textbf{Method} & \textbf{accuracy} & \textbf{fscore0} & \textbf{fscore1} & \textbf{fscore2} & \textbf{spearman rank} \\ 
 \hline\hline
    \multicolumn{6}{|l|}{Bert$_{large}$ + Naive Bayes(10)} \\
    \hline
    Pre-processing 1 &\textbf{0.46} & \textbf{0.60} & 0.02 & \textbf{0.28} & \textbf{0.288}  \\ 
    Pre-processing 2 & 0.46 & 0.60 & 0.02 & 0.28 & 0.267  \\ 
    Pre-processing 3 & 0.44 & 0.59 & \textbf{0.05} & 0.22 & 0.216  \\ 
 
    \hline
    \multicolumn{6}{|l|}{Bert$_{large}$ + Linear Regression(15)} \\
    \hline
    Pre-processing 1 & \textbf{0.48} & \textbf{0.50} & 0.28 & \textbf{0.60} & \textbf{0.398}  \\ 
    Pre-processing 2 & 0.48 & 0.48 & \textbf{0.29} & 0.60 & 0.363  \\ 
    Pre-processing 3 & 0.45 & 0.43 & 0.28 & 0.58 & 0.300  \\ 

 \hline
\end{tabular}
\caption{Compare Preprocessing Methods}
\label{table:T2}
\end{center}
\end{table*}

\begin{table*}[h]
\begin{center}
\begin{tabular}{|c|c|c|c|c|c|} 
 \hline
    \textbf{Method} & \textbf{accuracy} & \textbf{fscore0} & \textbf{fscore1} & \textbf{fscore2} & \textbf{spearman rank} \\ 
 \hline\hline
    \multicolumn{6}{|l|}{Naive Bayes} \\
    \hline
    Bert$_{large}$(10) & 0.46 & 0.60 & 0.02 & 0.28 & 0.289  \\ 
    N-gram(16) & 0.38 & 0.00 & 0.25 & 0.58 & 0.107  \\ 
    Electra & 0.39 & 0.01 & 0.25 & \textbf{0.58} & 0.088 \\
    Ensemble(10)+(16) & 0.46 & 0.60 & 0.02 & 0.29 & 0.290  \\ 
    Ensemble(10)+(16)+Electra& \textbf{0.46} & \textbf{0.60} & \textbf{0.02} & 0.29 & \textbf{0.295}  \\ 
 
    \hline
    \multicolumn{6}{|l|}{Linear Regression} \\
    \hline
    Bert$_{large}$(10) & \textbf{0.48} & \textbf{0.50} & \textbf{0.28} & \textbf{0.60} & \textbf{0.398}  \\ 
    N-gram(16) & 0.34 & 0.21 & 0.28 & 0.47 & 0.109  \\
    Electra & 0.36 & 0.38 & 0.22 & 0.44 & 0.151\\
    Ensemble(10)+(16) & 0.44 & 0.44 & 0.29 & 0.53 & 0.334  \\ 
    Ensemble(10)+(16)+Electra& 0.42 & 0.41 & 0.26 & 0.52 & 0.339  \\ 
 \hline
\end{tabular}
\caption{Compare Ensemble Methods}
\label{table:T3}
\end{center}
\end{table*}

\subsubsection{Using N-gram Probabilities}
 We use the previous word, the filling word, and the next word to form a tri-gram(or quadri-gram if the filling words contain 2 words). We acquire the n-gram frequency from Google Books N-gram model 1900 - 2019 and use it in replacement of the score mentioned in the previous section. We also applied similar classification and regression models. An additional rule is added to the regression model: if the n-gram frequency is 0, it's automatically mapped to negative:
 \begin{enumerate}
    \item{\textbf{Classification}:}
        \begin{enumerate}
            \item[(16)]{N-gram + Naive Bayes Classifier}
        \end{enumerate}
    \item{\textbf{Regression}: }
        \begin{enumerate}
            \item[(17)]{N-gram + Linear Regression}
        \end{enumerate}
 \end{enumerate}

 \subsubsection{Classification with replaced token detection (RTD) scores}
 As described in section 2.3 (ii) the Electra decoder takes a text sequence as input and for each token in the sequence outputs the probability that the token in corrupted/replaced. We use the RTD score corresponding to the filler word as input to the Naive Bayes classifier to get the predicted class for the (sentence, filler) pair.

\subsubsection{Domain adaption with MLM}
Although BERT aims to learn contextualized representations across a wide range of tasks, leveraging BERT alone still leaves the domain challenge and low resource challenges unsolved. BERT is trained on large text corpus(Wikipedia) and also requires a large number of samples to do fine-tuning. It suffers considerably at low-resource when applied to a different domain. However, in this task, we only have a limited number of samples(thousands of sentences), which is insufficient to fine-tune BERT to ensure full task awareness. So, we adopt BERT as the base model and adapt it to our specific domain using a masked language modeling(MLM) loss. 

We conduct domain adaption using the WikiHow dataset, a large-scale dataset using the online WikiHow knowledge base. It contains more than 200K articles, each of which is formed by merging multiple paragraphs.

\subsubsection{Pretraining with version distinction task}
Apart from pretraining the transformer models on the wikiHow dataset using MLM, we also try to pretrain using the version distinction task \cite{anthonio2020wikihowtoimprove} \cite{bhat2020towards}. The version distinction task involves differentiating between the older and newer version of a sentence in wikiHow. The dataset for this task was created using the revision history of the wikiHow corpus \cite{anthonio2020wikihowtoimprove}. Similar to the Bert Model discussed by the authors, we add a regression head to a transformer model to predict a score $s$ given an instruction. More formally, given a sentence pair $(s_i, s_j)$ corresponding to an old and new sentence, we first convert the sentences to numerical embeddings $(e_i, e_j)$ using a transformer encoder. Then the embeddings are passed to a stack of linear layers to obtain scores $(\phi(e_i), \phi(e_j))$. The transformer encoder and the linear layers are trained using the following pairwise ranking loss: 

$l_p = max(0, \phi(e_i) - \phi(e_j) + 1)$

We then train a simple Naive Bayes classifier using the 1-D scores from the above-described model to get the (sentence, filler) classification label.
We hypothesize that this model learns to score a sentence based on its grammatical and semantic correctness. As a result, for our cloze task, the model should give a higher score to plausible(sentence, filler) pairs compared to implausible pairs. Lastly, we use an Electra model as the transformer encoder as it is faster to train.

\subsubsection{Baselines}
Organisers of the task provided the following model as the baseline to improve on. 
\begin{itemize}
    \item [(0)]{tf-idf + Naive Bayes Classifier}
\end{itemize}

\subsection{Compare Preprocessing Methods}
Model (10) and Model (15) are used to evaluate different preprocessing methods(1, 2, and 3).

\subsection{Ensemble Experiments}
We use an ensemble score from (10) and (16) to train a classifier, and (15) and (17) to train a regression model, to see if there is an improvement compared to single models:

\section{Results}
From Table \ref{table:T1} under 'classification with sentence embeddings' we see that the Electra model has the best overall performance. In both Bert and XLNet the [MASK] token is only present during pre-training, which precedes the fine-tuning step. This results in different token distributions for the two stages. In contrast, the Electra decoder is trained on complete sequences with no masking. In the sentence embedding approach also, we replace the blank in the instruction with the filler and pass the complete sequence as input to the transformer. \

Further from Table  \ref{table:T1},  we observe that using MLM scores Bert outperforms both XLnet and Electra. The reason for this could be the bidirectional training of Bert. Further, as described in section 2.3 (ii) the noise in the input sequence to Electra discriminator is generated by a Masked Language model that is much smaller than the Bert base model. Therefore the corrupted sequences that the discriminator sees are not too complicated. As a result, the MLM score of Bert is more powerful than the RTD score to find the probability of a word given its context.\

By comparing regression models with classification models under the same pre-trained BERT model, we observed that regression models outperform classification models in total accuracy and predicting neutral cases while having high F1 scores for both positive and negative cases. The reason for this might be that regression takes the ordinal nature of the labels into account, and forced a balanced prediction at the same time. This total effect increased the model's performance in detecting neutral cases. 

For the 5 ways to get a score using MLM models in section 3.1.2, we discovered that the distance-based scores(model(11), (12), and (13)) perform poorly(30.6\%, 29.7\%, 32.9\% accuracy). And soft-max probability as a score performs slightly better than logit-based score. The reason behind this is unclear, but based on this empirical result, for other experiments, we decided to use the soft-max score-based models for MLM models. 

We also find that BERT$_{base+adapt}$ is slightly outperforming BERT$_{base}$ and close to BERT$_{large}$, especially in the accuracy and F1 score of different classes. This implies that domain adaption works, but due to the similarity of WikiHow and Wikipedia, the improvement is not huge. Similarly, pretraining using version distinction task also doesn't give high performance. The reason for this could be that a majority of the Wiki-Revisions dataset is characterized by the addition and deletion of tokens. Therefore the model does not see many cases similar to our downstream task where a particular token in instruction is replaced with tokens that can be lexically similar. \


From Table \ref{table:T2}, we find that concatenating titles, subheadings, and context instructions will improve the performance of pretrained language models(BERT). With context instructions having a larger impact compared to titles and subheadings. This indicates BERT can understand the long sequence dependencies, even when the constituent sub-sentences are from different parts of a document. \

Lastly, from Table \ref{table:T3}, Ensemble methods that combine raw prediction scores can further improve the classification performance on the best single model. To do so it disregards some underrepresented labels(Neural). For regression-based prediction models, the ensemble methods make balanced predictions on all possible classes. A possible explanation based on our experiment is: Bert captures the feature of the negative case well, while N-grams and Electra capture the features of the positive case well. A classification model is optimized towards higher prediction accuracy, thus considering the advantages of all models, while the shared disadvantage(underrepresented neutral class) remains. The regression model is designed to give balanced prediction results that are similar to the training set. Thus the underrepresented class is given more attention, at the cost of bringing down the prediction performance for other classes in the best models. This might result in worse performance compared to the best single model.

\section{Conclusion}

Our experiments are done in a manner that covers a broad range of models rather than exploiting a single model's performance with in-depth fine-tuning. Given the limited amount of training data, we deem it more meaningful to look at this task in the big picture and explore general directions to improve the performance of models for this task. In comparison to the baseline provided by the semeval task organizers, we improve accuracy by 10\%. For future work on a similar task, the biggest takeaways from our experiments are 1. Pre-trained transformers as off-the-shelf models do not give high performance in classification settings where the classes have ordinal nature and the difference in characteristics of the adjacent classes is small. 2. Ensemble predictions from multiple models can improve the performance for the well-represented classes(binary: good or bad), but not the underrepresented ones(neutral). 3. Given a small dataset(the size that's reasonable for a binary classification problem in language understanding), fine-tuning a pre-trained model cannot capture the standard of the labeler used to classify a neutral class.

\bibliography{acl2015}

\begin{thebibliography}{}

\bibitem[\protect\citename{Anthonio \bgroup et al.\egroup
  }2020]{anthonio2020wikihowtoimprove}
Talita Anthonio, Irshad Bhat, and Michael Roth.
\newblock 2020.
\newblock wikihowtoimprove: A resource and analyses on edits in instructional
  texts.
\newblock In {\em Proceedings of The 12th Language Resources and Evaluation
  Conference}, pages 5721--5729.

\bibitem[\protect\citename{Bhat \bgroup et al.\egroup }2020]{bhat2020towards}
Irshad Bhat, Talita Anthonio, and Michael Roth.
\newblock 2020.
\newblock Towards modeling revision requirements in wikihow instructions.
\newblock In {\em Proceedings of the 2020 Conference on Empirical Methods in
  Natural Language Processing (EMNLP)}, pages 8407--8414.

\bibitem[\protect\citename{Chang \bgroup et al.\egroup
  }2017]{chang2017weighted}
Chia-Yang Chang, Shie-Jue Lee, and Chih-Chin Lai.
\newblock 2017.
\newblock Weighted word2vec based on the distance of words.
\newblock In {\em 2017 International Conference on Machine Learning and
  Cybernetics (ICMLC)}, volume~2, pages 563--568. IEEE.

\bibitem[\protect\citename{Clark \bgroup et al.\egroup }2020]{clark2020electra}
Kevin Clark, Minh-Thang Luong, Quoc~V Le, and Christopher~D Manning.
\newblock 2020.
\newblock Electra: Pre-training text encoders as discriminators rather than
  generators.
\newblock {\em arXiv preprint arXiv:2003.10555}.

\bibitem[\protect\citename{Cui \bgroup et al.\egroup }2020]{cui2020does}
Leyang Cui, Sijie Cheng, Yu~Wu, and Yue Zhang.
\newblock 2020.
\newblock Does bert solve commonsense task via commonsense knowledge?
\newblock {\em arXiv preprint arXiv:2008.03945}.

\bibitem[\protect\citename{Devlin \bgroup et al.\egroup }2018]{devlin2018bert}
Jacob Devlin, Ming-Wei Chang, Kenton Lee, and Kristina Toutanova.
\newblock 2018.
\newblock Bert: Pre-training of deep bidirectional transformers for language
  understanding.
\newblock {\em arXiv preprint arXiv:1810.04805}.

\bibitem[\protect\citename{Hendrycks and Gimpel}2016]{hendrycks2016baseline}
Dan Hendrycks and Kevin Gimpel.
\newblock 2016.
\newblock A baseline for detecting misclassified and out-of-distribution
  examples in neural networks.
\newblock {\em arXiv preprint arXiv:1610.02136}.

\bibitem[\protect\citename{Lee \bgroup et al.\egroup }2017]{lee2017training}
Kimin Lee, Honglak Lee, Kibok Lee, and Jinwoo Shin.
\newblock 2017.
\newblock Training confidence-calibrated classifiers for detecting
  out-of-distribution samples.
\newblock {\em arXiv preprint arXiv:1711.09325}.

\bibitem[\protect\citename{Peters \bgroup et al.\egroup
  }2018]{peters-etal-2018-deep}
Matthew~E. Peters, Mark Neumann, Mohit Iyyer, Matt Gardner, Christopher Clark,
  Kenton Lee, and Luke Zettlemoyer.
\newblock 2018.
\newblock Deep contextualized word representations.
\newblock In {\em Proceedings of the 2018 Conference of the North {A}merican
  Chapter of the Association for Computational Linguistics: Human Language
  Technologies, Volume 1 (Long Papers)}, pages 2227--2237, New Orleans,
  Louisiana, June. Association for Computational Linguistics.

\bibitem[\protect\citename{Radford \bgroup et al.\egroup
  }2018]{radford2018improving}
Alec Radford, Karthik Narasimhan, Tim Salimans, and Ilya Sutskever.
\newblock 2018.
\newblock Improving language understanding by generative pre-training.

\bibitem[\protect\citename{Yang \bgroup et al.\egroup }2019]{yang2019xlnet}
Zhilin Yang, Zihang Dai, Yiming Yang, Jaime Carbonell, Russ~R Salakhutdinov,
  and Quoc~V Le.
\newblock 2019.
\newblock Xlnet: Generalized autoregressive pretraining for language
  understanding.
\newblock {\em Advances in neural information processing systems}, 32.

\end{thebibliography}
\bibliographystyle{acl}

Github\footnote{https://github.com/rchanumo/CS544\_project (Be careful of the underscore when copying)} and Video Presentation\footnote{https://www.youtube.com/watch?v=iWvDm0xILfw}

\end{document}